%% file: main.tex
\documentclass[letterpaper, 10 pt, conference]{ieeeconf}
\IEEEoverridecommandlockouts      
\usepackage[dvipsnames,table,xcdraw]{xcolor}
\usepackage[ruled,vlined,linesnumbered]{algorithm2e}
\usepackage{graphics} 
\usepackage{graphicx} %
\usepackage{lipsum,adjustbox}
\usepackage{tikz}
\usetikzlibrary{positioning, shapes.geometric}
\usepackage{verbatim}
\usetikzlibrary{calc}

\usepackage[caption=false,font=footnotesize]{subfig}
\usepackage{float}
\usepackage{amsmath} 
\usepackage{amssymb}  
\usepackage{mathtools}

\usepackage{mymath,gmech}
\usepackage{cite}
\let\labelindent\relax
\usepackage{enumitem}
\usepackage{mleftright}
\usepackage{booktabs}
\usepackage{cuted}
\usepackage{url}
\usepackage{balance}

\usepackage{multirow}
\graphicspath{ {../figs/}{../figures/}}

\newcommand{\rebuttal}[1]{\textcolor{black}{#1}}
\newcommand{\RALEdit}[1]{\textcolor{black}{#1}}

\makeatletter 
  \newcommand\figcaption{\def\@captype{figure}\caption} 
  \newcommand\tabcaption{\def\@captype{table}\caption} 
\makeatother

\title{\LARGE \bf Diffusion-Reinforcement Learning Hierarchical Motion Planning in Multi-agent Adversarial Games}

 \author{{Zixuan Wu$^{*}$, Sean Ye$^{*}$, Manisha Natarajan$^{*}$ and Matthew C. Gombolay$^{*}$}
 \thanks{$^{*}$Authors are from the Institute of Robotics and Intelligent Machines (IRIM), Georgia Institute of Technology, Atlanta, GA 30308, USA.}
\thanks{Correspondance Author: Zixuan Wu 
{\tt\small zwu380@gatech.edu}}
 }
\begin{document}

\maketitle
\thispagestyle{empty}
\pagestyle{empty}

\begin{abstract}
\rebuttal{Reinforcement Learning (RL)-based} motion planning has recently shown the potential to outperform traditional approaches from autonomous navigation to robot manipulation. In this work, we focus on a motion planning task for an evasive target in a partially observable multi-agent adversarial pursuit-evasion \rebuttal{game} (PEG). Pursuit-evasion problems are relevant to various applications, such as search and rescue operations and surveillance robots, where robots must effectively plan their actions to gather intelligence or accomplish mission tasks while avoiding detection or capture. 
We propose a hierarchical architecture that integrates a high-level diffusion model to plan global paths responsive to environment \rebuttal{data,} while a low-level RL policy reasons about evasive versus global path-following behavior. \rebuttal{The benchmark} results across different domains and different observability show \rebuttal{that} our approach \rebuttal{outperforms} baselines \RALEdit{by 77.18\% and 47.38\% on detection and goal reaching rate, which leads to 51.4\% increasing of the performance score on average.} Additionally, our method improves \RALEdit{interpretability, flexibility and efficiency of the learned policy}.\footnote{Code and supplementary details are at \url{https://github.com/ChampagneAndfragrance/Diffusion_RL}}


\end{abstract}


\input{intro.tex}

\input{lit_review}

\input{background.tex}

\input{method.tex}

\input{result.tex}
\input{conc.tex}



\balance
\bibliographystyle{IEEEtran}
\bibliography{MTS}


\end{document}

%% file: intro.tex
\section{Introduction \label{sec:Intro}}

Multi-agent pursuit-evasion games (PEG) \cite{vidal2002probabilistic} are ubiquitous in real-world robotics scenarios, where robotic systems could be applied to drug smuggler \rebuttal{tracking} \cite{silkman2001use}, defensive escorting \cite{escort}, or robot soccer \cite{robocup}. In PEG, one team seeks to track its target while the target agent or team employs evasive maneuvers, and research naturally focuses on developing methods for either the pursuers~\cite{afzalov2022strategy} or evaders~\cite{bulitko2006state, isaza2008cover, sigurdson2018multi}. \rebuttal{Although} there has been extensive work on developing search policies \cite{kennedy1995particle,liu2016topology,IMM1},  relatively little attention has been paid to evasion. This lack of emphasis provides an opportunity to design a novel evasive policy that is useful in real-world scenarios, such as business ships evading pirates \RALEdit{and evacuating lost people in national parks}.


\begin{figure}[t]
    \centering
    \includegraphics[width=\linewidth]{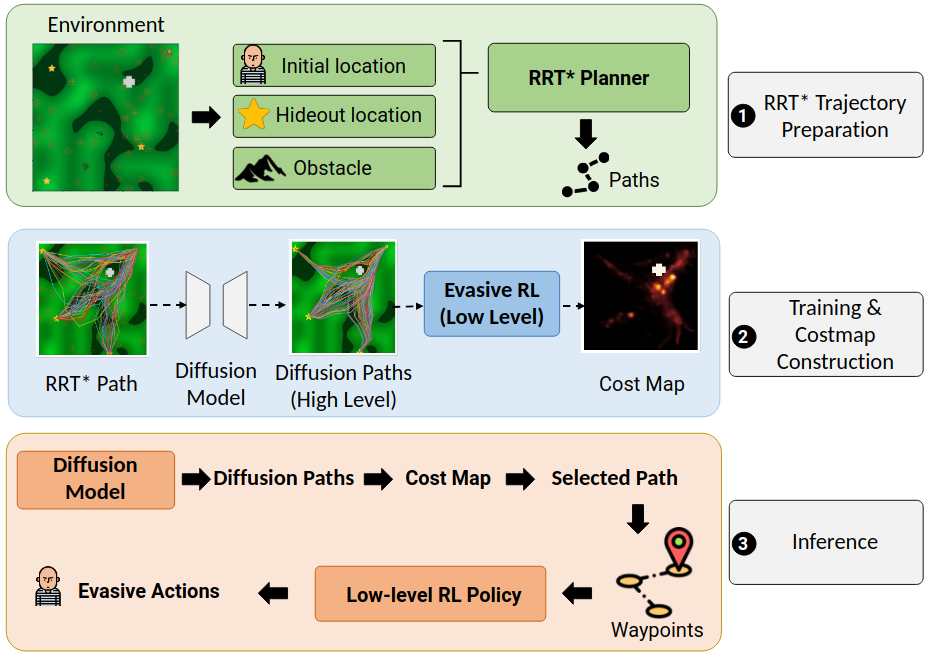}
    \vspace{-1.0em}
    \caption{Diffusion-RL Framework Overview: We first collect RRT* paths into a dataset. Then we use a diffusion model to learn the distribution of the RRT* path and generate samples as the high-level global plans to help learn a low-level evasive RL policy. A posterior costmap is built based on the learned hierarchy and detection risk and can be used to select the best global path in the inference stage.}
    \label{fig:diff_rl_overview}
    \vspace{-2em}
\end{figure}
Recently, learning-based methods, such as Graph Neural Networks \cite{ye2023learning}, diffusion models \cite{ye2023diffusion} or reinforcement learning (RL) methods \cite{wu2023adversarial} have been widely used in the pursuit-evasion games and similar navigation problems \cite{li2022hierarchical, liu2023dipper, carvalho2023motion} (e.g. goal-reaching, path-planning, etc.). These methods often outperform expert policies \cite{multiAUV}. However, learning the RL policy is usually subject to two key challenges---partial observability and high sample complexity when exploring large-scale, partially observable environments, as considered in our work.



Researchers propose hierarchical approaches to help exploration by structuring the goal, policies, or rewards. Goal-oriented reinforcement learning (for example, \rebuttal{a} replay of the hindsight experience \cite{andrychowicz2017hindsight}) increases the reward density by \rebuttal{assuming that} the goal has been reached in each exploration trajectory and \rebuttal{corrects} observation and reward accordingly. Hierarchical RL \cite{bacon2017option} improves sample efficiency from spatial or temporal abstraction by multiple hierarchically organized RL policies, and hybrid reward structure-based methods \cite{zhang2021distributional} try to capture the multimodal distribution of reward where each part of \rebuttal{the} value function can be a low-dimensional representation. However, their applications are tested in fully observable games \cite{bacon2017option} or require expert prior knowledge as primitives \cite{nasiriany2022augmenting}. 


In this paper, we propose a novel hierarchical learning framework that \rebuttal{combines} diffusion and RL as high- and low-level policies, respectively. The diffusion model is used as a global path generator to satisfy static map constraints, such as terminal states and obstacle avoidance. \rebuttal{Furthermore}, we employ a low-level RL policy that learns to follow the waypoints while prudently evading captures (see Figure \ref{fig:diff_rl_overview}). Our key insight is that using the diffusion model can improve the RL performance by constraining its exploration to high-value state-action regions and can help downstream tasks by providing diverse options. \textit{We find our method is especially useful for the evader to learn escaping behaviors in the large multi-agent, multi-goal partially observable settings --- Prisoner Escape and Narco Interdiction Domain (see \S\ref{sec:environment}).} 



\noindent \textbf{Contributions:} In summary, our contributions are three-fold:
\begin{enumerate}


    \item We propose a novel hierarchical system consisting of a diffusion model \RALEdit{as a high-level global path planner to aid RL exploration} and a low-level RL agent to learn evasive maneuvers, which outperforms all baselines.


    \item We design an algorithm to implicitly infer the cost map that is used to select paths within our hierarchical motion planning framework to evade the searching agents. \RALEdit{Multiple ablation studies show the interpretability, flexibility, efficiency, and generalizability of our method.}

    \item Our work is the first to \RALEdit{equip} evaders \RALEdit{with hierarchical motion planning} to learn escaping while reaching \rebuttal{the} navigation goal in large partially observable environments without expert primitives.
\end{enumerate}

%% file: lit_review.tex
\section{Related Work}


\subsection{Pursuit-Evasion Games} \label{sec:PEG_review}
A perfect information PEG (i.e., full observability) can be modeled as a differential game and solved using control theory \cite{salimi2020pursuit}. In comparison, the incomplete information PEG (i.e., partial observability) is a more realistic setting since agents usually do not have full knowledge of the environment at all time. Researchers commonly use probabilistic models \cite{feng2020uncertain, 1067989}
and formulate PEG as optimization problems to maximize the payoff \cite{wang2020cooperative, selvakumar2022min}. Therefore, recent research also focuses on \rebuttal{obtaining} the optimal policy \rebuttal{through} RL \cite{qu2023pursuit}. Unfortunately, most RL works focus on pursuit policies~\cite{zhou2022decentralized, de2021decentralized}. 
Some recent works \cite{yang2023large, baker2019emergent} have focused on evasive behaviors evolved from scratch, but these policies do not need to reach explicit navigation goals. In comparison, we derive an evader policy that can not only escape from pursuers but also reach goals in large, adversarial, and partially observable domains.

\subsection{Diffusion Models in Path Generating} \label{sec:diffusion_review}
Diffusion models are a class of generative models that have demonstrated impressive performance in generating sequential data \rebuttal{in} several applications, such as image synthesis and video generation \cite{dhariwal2021diffusion, ho2022cascaded, croitoru2023diffusion}. 
Recently, there have been several works employing diffusion models in trajectory generation tasks. Diffuser \cite{janner2022planning} trains a diffusion model for trajectory generation on offline datasets and plans future trajectories with guided sampling. \rebuttal{The diffusion policy} \rebuttal{extends} this work to imitation learning and \rebuttal{demonstrates} the ability to learn robust policies for various pushing tasks conditioned only on visual \rebuttal{input} \cite{chi2023diffusionpolicy}. Recent work \cite{ye2023diffusion} has extended these approaches for generating trajectories of multiple opponents. \RALEdit{In our work, we further use the diffusion model as a trajectory planner that allows the reinforcement learning agent to generate diverse plan options in our partially observable pursuit-evasion domain.}




\subsection{RL in Motion Planning} \label{sec:rl_review}
Recent work \rebuttal{has} demonstrated the success of RL for motion planning problems in dynamic environments \cite{gao2021vision, rubi2021quadrotor, cimurs2020goal, li2022hierarchical}.
Many papers discuss how to use RL to avoid obstacles by analyzing exteroceptive sensor information such as RGB cameras \cite{cimurs2020goal}, LIDARs \cite{rubi2021quadrotor, cimurs2020goal}, or \rebuttal{converted} laser data from images \cite{gao2021vision}. However, none of these \rebuttal{previous} works assumes an adversarial environment where the ``obstacles'' are intelligent pursuers, or the map is as large as in our scenario \RALEdit{with undetectable parts}. 

The work \cite{li2022hierarchical} also employs a hierarchical structure in goal-conditioned planning\rebuttal{;} however, their environment does not incorporate adversaries or multiple goals\rebuttal{,} and their RL policy is not trained from scratch. Dynamic obstacles are considered in \cite{kastner2021arena}; however, these obstacles can only run on a \rebuttal{predefined} segment with a constant velocity, and \RALEdit{the map is known} such that A* search can be \RALEdit{sufficient to find the optimal path}. In our scenario, we assume the cameras are \RALEdit{hidden from} the evader \RALEdit{in a non-discovered terrain map}, while the pursuit team can \rebuttal{use} these cameras to detect and track the evader. 



%% file: background.tex
\newcommand{\roboF}{g}
\newcommand{\roboBV}{\xi_u}
\newcommand{\camF}{h}
\newcommand{\camBV}{\zeta_u}
\newcommand{\camProj}{H}
\newcommand{\camPt}{r}

\newcommand{\imJac}{{L}}
\newcommand{\imJacV}{\boldsymbol{L}}

\newcommand{\geoJac}{{G}}

\newcommand{\numFeat}{n_{\text{F}}}

\newcommand{\pose}{\boldsymbol{P}}
\newcommand{\control}{\boldsymbol{u}}
\newcommand{\transCR}{^C\boldsymbol{T}_{R}}
\newcommand{\transCRv}{^C\boldsymbol{T}_{R,v}}
\newcommand{\transCRw}{^C\boldsymbol{T}_{R,\omega}}

\newcommand{\pointSet}{\boldsymbol{Q}}
\newcommand{\pointSetV}{\boldsymbol{q}}

\newcommand{\featureSet}{\boldsymbol{S}}
\newcommand{\featSetV}{\boldsymbol{s}}
\newcommand{\featSetDV}{\boldsymbol{s}^*}
\newcommand{\featSetDVdot}{\dot{\boldsymbol{s}}^*}
\newcommand{\featureSetA}{\boldsymbol{S}_a}
\newcommand{\featureSetD}{\boldsymbol{S}^*}
\newcommand{\featureSetDA}{\boldsymbol{S}_a^*}

\newcommand{\error}{\boldsymbol{e}}
\newcommand{\errorSet}{\boldsymbol{E}}

\newcommand{\imageJacobian}{\boldsymbol{L}}

\newcommand{\imageJacobiani}{\boldsymbol{L}_{S_i}}
\newcommand{\imageJacobianv}{\boldsymbol{L}_{S,v}}
\newcommand{\imageJacobianw}{\boldsymbol{L}_{S,\omega}}
\newcommand{\transFeature}{g}
\newcommand{\homo}{H}

\section{Environment}
\begin{figure*}[t]
    \centering
    \includegraphics[width=0.3\linewidth,height = 4.5 cm]{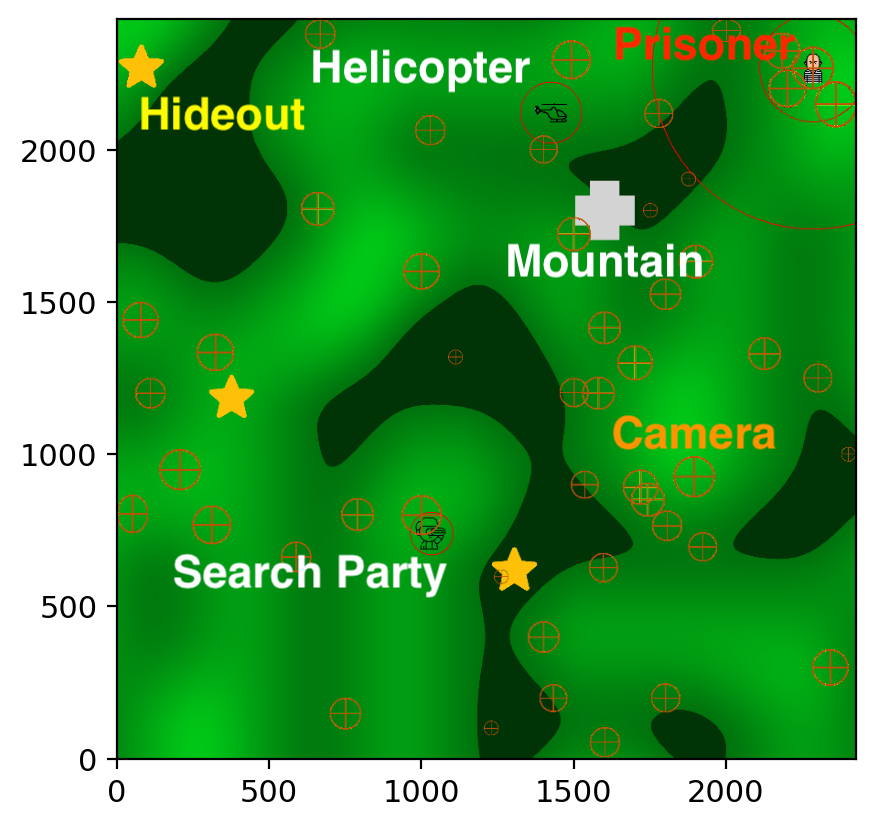}
    \includegraphics[width=0.6\linewidth,height = 4.5 cm]{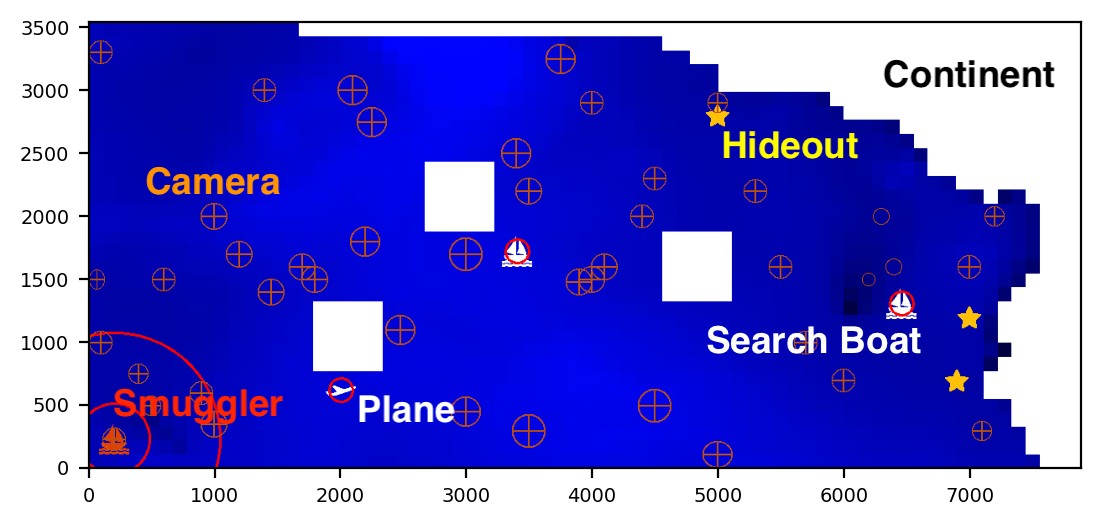}
    \vspace{-1.0em}
    \caption{Prisoner Escape Domain (left) and Narco Traffic Interdiction Domain (right)}
    \label{fig:escape}
    \vspace{-1.5em}
\end{figure*}
\subsection{Partially Observable Markov Decision Process}
We model adversarial search and tracking as a Partially Observable Markov Decision Process (POMDP). A POMDP for an agent can be defined by a state $s$, private observation $o$, action $a$, \rebuttal{and} a transition function $\mathcal{T}: s \times a \mapsto s$. At each timestep $t$, an agent receives an observation $o^t$, chooses an action $a^t \in \mathcal{A}$, and obtains a reward $r^t: s \times a_i \mapsto \mathbb{R}$. The initial world state is drawn from a prior distribution $\rho$. In our environment (see \S\ref{sec:environment}), we assume \rebuttal{that} the evader starts from a random position \rebuttal{from} a corner region of the map while the pursuit team starts from random positions on the map. \RALEdit{It is a partially observable environment for the evader in that 1) cameras are invisible to the evader, 2)  the terrain map is not provided a prior, and 3) the evader may only have \rebuttal{a} limited detection range.}


\subsection{Prisoner Escape and Narco Interdiction}
\label{sec:environment}

In our evaluation of the proposed algorithm, we employ two \RALEdit{open-sourced} large-scale pursuit-evasion domains \cite{ye2023learning} named Prisoner Escape and Narco Interdiction (Figure \ref{fig:escape}). In both domains, an evader (prisoner or smuggler) is being tracked by a team of search agents in a partially observable environment. \rebuttal{In each episode, three hideouts are drawn from all candidate hideouts.} These hideouts are only visible to the evader and unknown to the pursuit team\rebuttal{, as} adversarial teams do not communicate intent. The terminal objective of the evader is to reach any hideout on the map (considered a success) or reach the time limit (considered a failure). \RALEdit{Simultaneously, the evader should learn to avoid pursuit team detection. }The heterogeneous pursuit team consists of static (cameras) and dynamic agents (search parties, helicopters, search boats, and planes) that collaborate to search and track the evader in both domains. 


The maximum speed of the evader is always lower than the pursuit agents, making it more challenging for the evader to plan evasive maneuvers to successfully reach the hideouts undetected. The observation of the evader includes the current timestep, hideout locations, mountain locations, pursuer states, and evader states. \RALEdit{The evader can see the pursuer anywhere in the global-view mode and within a field-of-view in the local-view mode}, and the camera locations and visibility map are always hidden from the evader. \RALEdit{In the local-view mode, evader's field-of-view, $d_e$, is parameterized by the searching agent type, $\beta_{type}$, visibility, $c_v$, and searching agent speed, $s_s$: $d_e = \alpha\cdot(\beta_{type}\cdot c_v\cdot s_s + \eta)$.} These domains increase the complexity of the task and more closely match real-world dynamics compared to other pursuit-evasion games \cite{hungerlander2014discretetime,  sani2020pursuit, salimi2020pursuit}.

\subsection{Pursuit Team Behavior}
\label{sec:heuristic}
\RALEdit{To provide a fair comparison between our evading algorithm and the baselines, }we use a fixed heuristic policy for the pursuit team inspired by prior works \cite{spiralLawn, interception} that \RALEdit{achieved good performance in the prisoner escape domain \cite{wu2023adversarial}}. The search agents can choose to either 1) go to the last detection, 2) \RALEdit{search along the predicted path}, or 3) search in the vicinity of the last detection. Searching team detection range, $d_s$, is parameterized by the agent type, $\beta_{type}$, visibility, $c_v$, and evader speed $s_e$: $d_s = \alpha\cdot(\beta_{type}\cdot c_v\cdot s_e + \eta)$, which means a higher visibility and evader speed will lead to larger detection range. The pursuer policy is designed to effectively utilize the available information and adapt to different scenarios.

%% file: method.tex
\newcommand{\mat}{\boldsymbol}

\section{Method}\label{sec:method}
The evader's objective is to reach one of the hideouts on the map and evade detection by the pursuit team, using only partial map knowledge, without any information about camera locations or models for whether the pursuers can detect the evader. RL or diffusion-based motion planning each partially fits our setting: RL can hypothetically learn to evade through \rebuttal{trial-and-error}. However, RL has difficulties with exploration in the large, adversarial, and multi-goal domains. On the other hand, diffusion-based motion planning can generate diverse plans that reach the goals; however, diffusion models struggle to learn evasive behaviors and avoid the pursuing agents. The reason is that the diffusion model has a relatively long denoising process that is difficult to handle environment dynamics. Additionally, partial observability induces high variance in training.


We propose to leverage the strengths of diffusion and RL policies by using diffusion models to guide RL exploration during training and generate path candidates at inference. We decompose the hierarchy into the following: learning diffusion-based global plans that meet start, terminal, and obstacle constraints from RRT* paths (\S\ref{sec:diffusion}), employing RL local evasive behaviors to avoid detection en-route (\S\ref{sec:sac}), and path selection from posteriorly constructed costmap during inference (\S\ref{sec:map}). 



\SetCommentSty{mybluecomment}
\newcommand{\mybluecomment}[1]{\textcolor{blue}{\small\textit{#1}}}
\SetKwComment{Comment}{$\lhd$ }{}
\begin{algorithm}[hbt!]
\caption{Diffusion Model Training}\label{algo:diffusion_train}
\KwData{Start Location $\mathcal{S}$, Goal Locations $\mathcal{G}$, Obstacle Location $\mathcal{M}$, Constraints $\mathcal{C}=<\mathcal{S}, \mathcal{G}, \mathcal{M}>$, Data buffer $\mathcal{D}$, Noise Scheduling Terms $\alpha_i, \bar{\alpha_i}$, Learning Rate $\alpha$, Denoising Timesteps $T$}
\KwResult{Diffusion Model $s_\theta$}
\While{buffer $\mathcal{D}$ not full} 
{

$c=<p_s, p_g, m> \sim \mathcal{C}$ \Comment*[r]{Draw global constraints}
$\tau \gets RRT^{*}(c)$ \Comment*[r]{Plan RRT* path} \label{line:RRT_path}
$\tau \gets Sample_{\downarrow}(\tau)$ \Comment*[r]{Uniformly downsample path} \label{line:RRT_downsample}
$\mathcal{D} \gets \tau, c$ \Comment*[r]{Store trajectory, constrains in dataset}
}
\While{not converged}{ \label{line:train_diffusion}
\Comment{Draw $\tau$, $c$ diffusion timestep, and random noise}
$\tau_0, c \sim \mathcal{D}$\, $i \sim \mathcal{U}(1, T), \epsilon \sim \mathcal{N}(0, I)$\; \label{line:sample_2}
\Comment{Compute Score Matching MSE Loss} 
$\tau_i \leftarrow \sqrt{\bar{\alpha}_i} \tau_0 + \sqrt{1 - \bar{\alpha}}_i \epsilon$\;
$\mathcal{L}(\theta) = \| \epsilon - s_\theta(\tau_i, i, c))\|_2^2$\; \label{line:compute_loss}
\Comment{Update Model}
$\theta = \theta + \alpha \nabla_\theta \mathcal{L}(\theta) $ \label{line:update_params} 
}
\end{algorithm}
\begin{algorithm}[hbt!]
\caption{Diffusion Path Generator $\pi^d$}\label{algo:diff_sampling}
\KwData{Constraints $\mathcal{C}$}
\KwResult{Global plan $\tau^0$}
\While{$\tau^0$ not satisfies constraints $\mathcal{C}$}{
$\tau^T \leftarrow$ sample from $\mathcal{N}(0, I)$ \Comment*[r]{Gaussian Noise} \label{line:noise}
\For{all $i$ from $T$ to $1$}{ \label{line:for}
$(\mu^i, \Sigma^i) \leftarrow s_{\theta}(\tau^{i}), \Sigma_{\theta}(\tau^{i})$ \\ \label{line:model}
     $\tau^{i-1} \sim \mathcal{N}(\mu^i, \Sigma^i)$ \Comment*[r]{Diffusion Denoise} \label{line:sample}
     $\tau^{i-1} \gets \mathcal{C}(\tau^{i-1})$ \Comment*[r]{Apply Constraints} \label{line:constraints}
 }
 }
\textbf{Return} $\tau^0$
\end{algorithm}
\subsection{Diffusion Based Global Plan}\label{sec:diffusion}
In this section, we describe how we leverage diffusion models as a generative motion planner to provide a prior for the RL agent to traverse in large, partially observable, and adversarial environments \cite{10416776}. We utilize a diffusion model to \RALEdit{clone the distribution from RRT* planner} and \RALEdit{generate waypoints of diverse candidate paths} that the agent should follow approximately. This diffusion path planner guides the exploration, increases the sample-time efficiency, \RALEdit{and improves the flexibility} of the downstream RL algorithm. \textit{The underlying insight is to constrain the RL exploration at a promising distribution implicitly encoded by the diffusion model considering the map constraints.}


The diffusion path planning should have three key attributes: 1) global constraints are satisfied (start and terminal locations, obstacles, etc.);  2) generated plans are diverse and multimodal and 3) the sampling time of the diffusion model is low. We guide the diffusion sampling procedure using constraint guidance \cite{ye2023diffusion} to best satisfy the constraints and use a small number of sampling timesteps to keep the sampling time low. The low sampling time allows us to run the diffusion model during the RL training procedure, which is difficult for traditional \rebuttal{search-based} planners. Additionally, the parallel sampling \RALEdit{reduces the overhead} at inference time and provides diverse candidates for the agent to use. 


We outline the diffusion training process in Algorithm \ref{algo:diffusion_train}. We first generate paths with RRT* (Line \ref{line:RRT_path}) --- a sample-based motion planner that yields varied paths even for the same initial and target states, which enables the diffusion model to similarly generate diverse paths. Then we uniformly downsample the path into sparse waypoints (Line \ref{line:RRT_downsample}) and train a diffusion model based on these waypoints (Line \ref{line:train_diffusion}-\ref{line:update_params}). Generating a path with a few waypoints is sufficient enough to act as the \rebuttal{high-level} planner as we expect the RL policy to act as the low-level dynamic actor to move between waypoints. Generating fewer waypoints also allows sampling time to be faster, which is crucial for training in an RL loop. 


The sampling process is summarized in Algorithm \ref{algo:diff_sampling} with diffusion path generator, $\pi^{d}$. We apply global constraints at each denoising step (Line \ref{line:constraints}) to help generate a valid path from start to goal. 

\begin{algorithm}[hbt!]
\caption{Adversarial Behavior Learning}\label{alg:two}
\KwData{Initial state $s_0$, Diffusion Global Planner $\pi^d$, SAC replay buffer $\mathcal{D}_s$}
\KwResult{SAC low-level policy $\pi^{l*}$}
\While{training episodes}{
$\{w_i\}_{i=0}^{N_w}=\mathbf{w} \sim \pi^{d}(\mathbf{w}\vert s_0)$ \Comment*[r]{Waypoints from $\pi^d$} \label{line:waypt}
\While{$w_i$ not reached}{
$O_{aug} \gets [O\vert w_i]$  \Comment*[r]{Concat current waypoint} \label{line:augment}
$a\sim \pi^{l}(a\vert O_{aug})$ \Comment*[r]{Action from low policy} 
$O_{aug}^{\prime}, r \gets \mathbf{EnvStep}(a)$ \Comment*[r]{Update env} 
$\mathcal{D}_s \gets \{O_{aug}, a, O_{aug}^{\prime}, r\}$ \Comment*[r]{Push to buffer}
\If{$w_i$ reached}
{
$i \gets i+1$ \Comment*[r]{Update to next waypoint} 
}
}
$\pi^{l*} \gets SAC(d_s\sim\mathcal{D}_s)$ \Comment*[r]{Update SAC parameters} 
}
\end{algorithm}
\subsection{RL Based Adversarial Behavior Learning}\label{sec:sac}
In addition to our diffusion model considering the global objectives, it is also essential to design an algorithm directly interacting with the environment that can not only accomplish the tasks from the high-level planners but also capture the dynamic patterns in the map to achieve the local adversarial detection avoiding goals. Although evading heuristics exist  \cite{ye2023learning}, we believe that a data-driven reinforcement learning algorithm is a more convenient way compared with the hand-crafted policy and can perform better within specific environment distribution.

Therefore, we employ an RL algorithm Soft Actor-Critic (SAC) as the low-level policy with the following aims: 1) approximately follow the waypoints from the diffusion path and 2) adapt the evader velocities to escape from searching agents. We summarize RL training in Algorithm \ref{alg:two} where we augment the observation with the current desired waypoint from the high-level planner (Line \ref{line:waypt}, \ref{line:augment}). The training objective can be formulated as Equation \eqref{eq:diffusion_rl} where $\approx$ means \rebuttal{a similar} low-level policy can be derived from both sides.

{\small
    \begin{align}
    R=\max_{\pi}\mathbb{E}_{\pi}&\left[\sum_t \gamma^t \cdot r \right] \nonumber\\
    \approx\max_{\pi^{g}}\mathbb{E}_{\pi^{g}}&\left[\max_{\pi^{l}}\mathbb{E}_{\pi^{l}}\left[ \sum_{\tau=0}^{t_g}\sum_{t=0}^{t_l} \gamma^{t} \cdot r \bigg| w^{\tau}\sim\pi^{g}(w\vert s^{\tau}) \right]\right] \nonumber \\
    \approx\max_{\pi^{l}}\mathbb{E}_{\pi^{l}}&\left[ \sum_{\tau=0}^{t_g}\sum_{t=0}^{t_l} \gamma^{t} \cdot r \bigg| \mathbf{w}\sim\pi^{d}(\mathbf{w}\vert s_0), s_0\sim\rho \right] \label{eq:diffusion_rl}
    \end{align}
}
\noindent$\rho$ is the initial environment distribution for the start and hideout locations  ($s_0\sim\rho$ is omitted from the derivation), $\pi^{d}$ is the diffusion global planner, and $\mathbf{w}$ is the generated waypoints to be tracked, $t_g$ is the number of waypoints. Equation (\ref{eq:diffusion_rl}) shows that we \rebuttal{can} decompose our adversarial evading learning problem into a sequential optimization problem, including a waypoint generator and a low-level waypoint tracker, both of which aim to maximize the discounted accumulative reward. However, the unconstrained random exploration of these policies will \rebuttal{be} subject to high variance from the partial observability; therefore, we confine our exploration in a more reasonable region described by the diffusion policy distribution in the previous section. We set reward terms to guide the evader to follow the waypoint and avoid detection in Equation \eqref{eq:reward} where $r_g$ is the reward obtained when reaching the waypoint, \RALEdit{$r_{adv}$ is the detection penalty}, and $r_d$ is the distance penalty to the next waypoint.
    \begin{equation} \label{eq:reward}
        r = r_g + r_d + r_{adv}
    \end{equation}

\begin{table*}[]
\centering
\caption{Diffusion-RL benchmarks (mean $\pm$ std), G: global-view mode, L: local-view mode}
\label{tab:diffusion_rl_metrics}
\resizebox{\textwidth}{!}{%
\begin{tabular}{lccccccclccccccc}
\hline
\textbf{Domain}                                       & \multicolumn{3}{c}{\textbf{Prisoner Escape (G)}}                      & \textbf{} & \multicolumn{3}{c}{\textbf{Narco Interdiction (G)}}                   &           & \multicolumn{3}{c}{\textbf{Prisoner Escape (L)}}                      & \textbf{} & \multicolumn{3}{c}{\textbf{Narco Interdiction (L)}}                   \\ \hline
                                                      & Score $\uparrow$ & Detection $\downarrow$ & Goal-Reach $\uparrow$ &           & Score $\uparrow$ & Detection $\downarrow$ & Goal-Reach $\uparrow$ & \textbf{} & Score $\uparrow$ & Detection $\downarrow$ & Goal-Reach $\uparrow$ &           & Score $\uparrow$ & Detection $\downarrow$ & Goal-Reach $\uparrow$ \\ \hline
\multicolumn{1}{l|}{\textbf{Non-Learning Approaches}} &                  &                        &                       &           &                  &                        &                       &           &                  &                        &                       &           &                  &                        &                       \\
\multicolumn{1}{l|}{A-Star Heuristic}                 & 0.72$\pm$0.19    & 0.34$\pm$0.24          & 0.98$\pm$0.14         &           & 0.64$\pm$0.30    & 0.39$\pm$0.31          & 0.80$\pm$0.40         &           & 0.80$\pm$0.18    & 0.25$\pm$0.15          & 1.00$\pm$0.00         &           & 0.53$\pm$0.29    & 0.51$\pm$0.31          & 0.76$\pm$0.43         \\
\multicolumn{1}{l|}{RRT-Star Heuristic}               & 0.73$\pm$0.18    & 0.32$\pm$0.22          & 0.94$\pm$0.24         &           & 0.66$\pm$0.31    & 0.38$\pm$0.32          & 0.83$\pm$0.38         &           & 0.77$\pm$0.11    & 0.29$\pm$0.14          & 1.00$\pm$0.00         &           & 0.59$\pm$0.26    & 0.46$\pm$0.30          & 0.84$\pm$0.37         \\
\multicolumn{1}{l|}{VO-Heuristic}                     & 0.74$\pm$0.11    & 0.32$\pm$0.14          & 0.98$\pm$0.14         &           & 0.64$\pm$0.10    & 0.43$\pm$0.12          & 1.00$\pm$0.00         &           & 0.73$\pm$0.12    & 0.33$\pm$0.15          & 0.99$\pm$0.10         &           & 0.64$\pm$0.10    & 0.43$\pm$0.11          & 1.00$\pm$0.00         \\
\multicolumn{1}{l|}{\textbf{Learning Approaches}}     &                  &                        &                       &           &                  &                        &                       &           &                  &                        &                       &           &                  &                        &                       \\
\multicolumn{1}{l|}{DDPG}                             & 0.23$\pm$0.34    & 0.73$\pm$0.40          & 0.07$\pm$0.26         &           & 0.31$\pm$0.32    & 0.65$\pm$0.37          & 0.00$\pm$0.00         &           & 0.23$\pm$0.34    & 0.73$\pm$0.40          & 0.07$\pm$0.26         &           & 0.26$\pm$0.36    & 0.70$\pm$0.42          & 0.00$\pm$0.00         \\
\multicolumn{1}{l|}{SAC}                              & 0.78$\pm$0.12    & 0.06$\pm$0.11          & 0.13$\pm$0.34         &           & 0.49$\pm$0.35    & 0.43$\pm$0.41          & 0.01$\pm$0.10         &           & 0.73$\pm$0.12    & 0.10$\pm$0.15          & 0.05$\pm$0.22         &           & 0.51$\pm$0.40    & 0.41$\pm$0.47          & 0.00$\pm$0.00         \\
\multicolumn{1}{l|}{Diffusion Only}                   & 0.78$\pm$0.09    & 0.28$\pm$0.12          & 1.00$\pm$0.00         &           & 0.78$\pm$0.13    & 0.26$\pm$0.15          & 1.00$\pm$0.00         &           & 0.79$\pm$0.10    & 0.26$\pm$0.13          & 1.00$\pm$0.00         &           & 0.78$\pm$0.13    & 0.26$\pm$0.15          & 1.00$\pm$0.00         \\
\multicolumn{1}{l|}{\textbf{Our Approaches}}          &                  &                        &                       &           &                  &                        &                       &           &                  &                        &                       &           &                  &                        &                       \\
\multicolumn{1}{l|}{Diffusion-RL}                     & 0.90$\pm$0.09    & 0.12$\pm$0.10          & 0.96$\pm$0.20         &           & 0.89$\pm$0.11    & 0.12$\pm$0.10          & 0.94$\pm$0.24         &           & 0.90$\pm$0.12    & 0.11$\pm$0.10          & 0.90$\pm$0.30         &           & 0.89$\pm$0.11    & 0.11$\pm$0.11          & 0.91$\pm$0.29         \\
\multicolumn{1}{l|}{Diffusion-RL-Map}                 & \textbf{0.94$\pm$0.08}    & 0.07$\pm$0.08          & 0.98$\pm$0.14         &           & \textbf{0.94$\pm$0.07}    & 0.07$\pm$0.08          & 1.00$\pm$0.00         &           & \textbf{0.95$\pm$0.09}    & 0.05$\pm$0.07          & 0.95$\pm$0.22         &           & \textbf{0.94$\pm$0.08}    & 0.06$\pm$0.07          & 0.95$\pm$0.22         \\ \hline
\end{tabular}%
}
\vspace{-1.5em}
\end{table*}

\subsection{Cost Map Construction and Inference}\label{sec:map}
\RALEdit{Having discussed how the diffusion path planner could help the RL explore within the reward promising region in \S\ref{sec:sac} and how the low-level SAC policy could be derived under the global path and initial environment distribution, we now take a step further and describe which global path sample should be drawn in each test case.}

Even though \rebuttal{the} evasive agent does not have direct access to the camera or the terrain map information,
\RALEdit{this agent can still perform reverse spatial inference on the map by interacting with the learned policies \rebuttal{to} identify regions of heightened risk. Following this idea, we construct a cost map to reflect the risk of detection which we can then use to score the global paths.} 



Algorithm \ref{alg:costmap} describes our procedure. We run a certain number of episodes after the hierarchy is trained well (Line \ref{line:rollout}) and simultaneously record the evader location $\mathbf{X}$ and its distance to the closest searching agent $\mathbf{d}$ (Line \ref{line:record}). \RALEdit{If the searching agent is within the evader's field of view,} we select out the evader locations $x_i$ with which the associated distance, $d_i$, is below a risk threshold $\epsilon$ and add a Gaussian distribution whose mean is $x_i$ and standard deviation $\sigma$ (Line \ref{line:sel_loc}-\ref{line:gaussian}). Next, we perform normalization and output the map. 

Importantly, end-users can preview the global paths before execution or revise the cost map with the newly added global information, which adds more flexibility and interpretability compared with traditional fully \rebuttal{MDP-based} RL algorithms (Line \ref{line:normalize}). We can select the global path with \rebuttal{the} lowest cost based on the costmap (Line \ref{line:inference_start}-\ref{line:inference_end}). 


\begin{algorithm}[hbt!]
\caption{Costmap and Inference}\label{alg:costmap}
\KwData{Initial state $s_0$, Diffusion Global Planner $\pi^d$, Learned local policy $\pi^{l*}$, Costmap $\mathbf{G}$, Evasive agent location $\mathbf{x}$}
\KwResult{Selected global plan $\mathbf{w}^{*}$}
\While{costmap constructing episodes}{ \label{line:rollout}
\Comment{Continue running episodes with learned policies} 
$\mathbf{x}, \mathbf{d} = \mathbf{Env}(s_0, \pi^d, \pi^{l*})$ \\ \label{line:record}
\For{all $x_i, d_i$ from $\mathbf{x},\mathbf{d}$}{ \label{line:sel_loc}
\Comment{Update costmap when search team is close} 
\If{$d_i<\epsilon$ \& see searching agent} 
{
$\mathbf{G} \gets \mathbf{G} + \mathcal{N}(x_i, \sigma^{2})$ \Comment*[r]{Add a Gaussian} \label{line:gaussian}
}
}
\Comment{Normalize and adjust costmap} 
$\mathbf{G} \gets adjust(\mathbf{G}/\mathbf{G}_{max})$\; \label{line:normalize}
\Comment{Get global path samples} 
$\mathcal{T} \gets \{\mathbf{w}_i: \mathbf{w}_i \sim \pi^{d}(\mathbf{w}\vert s_0)\}_{i=0}^{N_s}$ \; \label{line:inference_start}
\Comment{Find global path with lowest cost} 
$\mathbf{w}^{*}=\argmax_{\mathbf{w}_i\in\mathcal{T}}\int_{\mathbf{w_i}} \mathbf{G}(w)dw$ \;\label{line:inference_end}
}
\end{algorithm}

%% file: result.tex
\section{Results \& Discussion}\label{sec:exp}

In this section, we will first benchmark our diffusion-RL framework \RALEdit{for the global- and local-view evader} against both heuristics and learning based baselines (see \S\ref{sec:challenge}). Then we visualize our constructed costmap and show \RALEdit{the interpretability, efficiency, and generalizability} of our method \RALEdit{in multiple ablation studies (see \S\ref{sec:costmap_res}-\S\ref{sec:generalizability})}. Next, we validate our hierarchy in the real robot testbed (see \S\ref{sec:robotarium}). Finally, we include a result discussion section (see \S\ref{sec:discussion}).



\begin{figure}[h]
    \centering
    \includegraphics[width=0.7\linewidth]{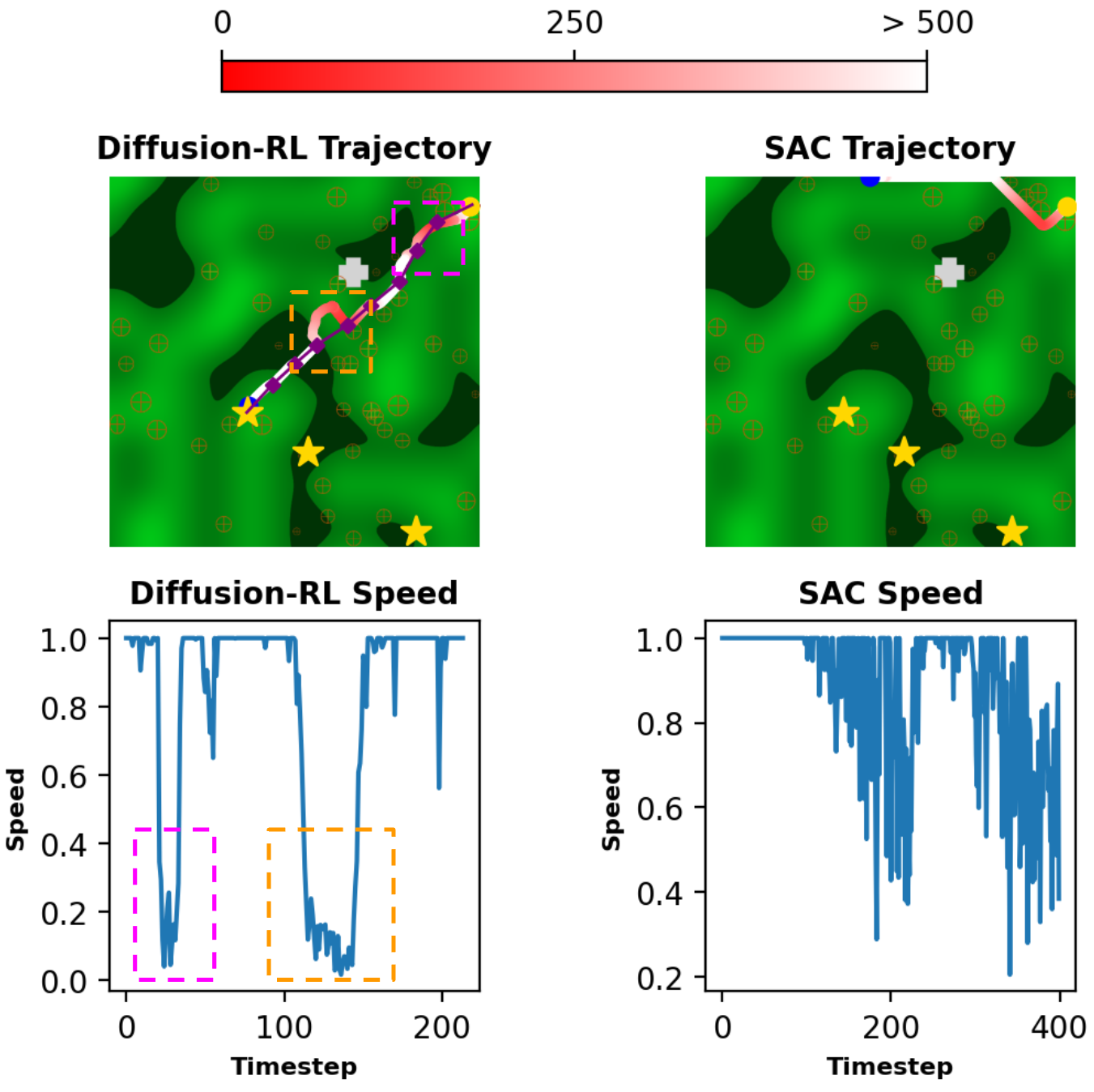}
    \vspace{-1em}
    \caption{Our Diffusion-RL trajectory vs SAC trajectory. Colorbar indicates \rebuttal{the distance} between the evader and closest pursuer and the purple line indicates the diffusion global path.}
    \label{fig:diff_vs_sac}
    \vspace{-1.5em}
\end{figure}
\begin{figure*}[h]
    \centering
    \subfloat[Costmap]{\includegraphics[width=0.18\textwidth]{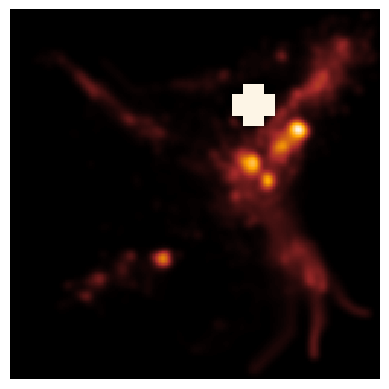} \label{fig:costmap}}
    \hspace{0.5em}
    \subfloat[Camera Locations]{\includegraphics[width=0.18\textwidth]{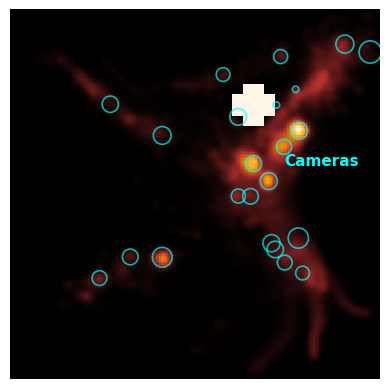} \label{fig:cam}}
    \hspace{0.5em}
        \subfloat[Selected Path]{\includegraphics[width=0.18\textwidth]{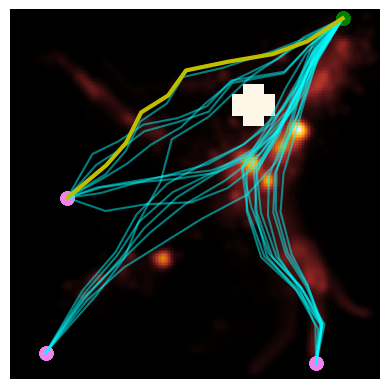}\label{fig:selected_path}}
    \hspace{0.5em}
        \subfloat[Ad Hoc Map]{\includegraphics[width=0.18\textwidth]{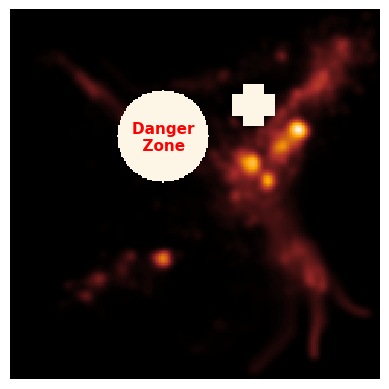} \label{fig:new_zone}}
    \hspace{0.5em}
        \subfloat[New Selected Path]{\includegraphics[width=0.18\textwidth]{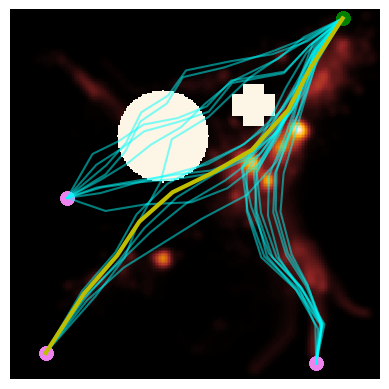} \label{fig:new_path}}
    \vspace{-0.5em}
    \caption{Path Planning with the Costmap: The costmap (a) is constructed by correlating the agent's risk of detection to its location on the map. We show in (b) that the agent \rebuttal{can successfully} identify where the cameras are. Given this costmap, the agent can select a path that best evades \rebuttal{high-cost} regions (c). Additional obstacles can be added ad-hoc (d) and a new path can be chosen (e). The grey areas indicate untraversable obstacles or danger zones.}
    \label{fig:hier}
    \vspace{-1.0em}
\end{figure*}
\subsection{Achieving Long-Term Planning and Short-Term Evasion}\label{sec:challenge}
We propose to use the following metrics to quantitatively evaluate our method and compare it against the baselines: detection, goal-reaching rate, and score ($\uparrow$ means higher is better and $\downarrow$ means lower is better in Table \ref{tab:diffusion_rl_metrics}). Detection is the average number of steps when the evader is detected for each episode. Goal-reaching rate is the number of episodes in which the evader reaches the goal within the time limit divided by the total number of episodes tested. The score metric is a weighted reward term that balances the two main objectives in our domains: reaching the goal ($\pm$50) and staying undetected by the opposing blue agents (-1 for each detection). Each metric is normalized by the min-max value of each category and we test the algorithms in 100 episodes with different random seeds from the training.

We compare our diffusion+RL method with learning based and \rebuttal{non-learning-based} baselines. A* and RRT* heuristics are from the \RALEdit{open-sourced environment} \cite{ye2023learning} where the evader follows the A* or RRT* path and \rebuttal{shows} some pre-scripted adversarial behaviors (e.g. heading to dense forest, changing velocity, etc.) when the search team is nearby. Velocity obstacle (VO) method is widely used in dynamic obstacle avoidance tasks which samples a velocity from the \rebuttal{collision-free} velocity space \cite{fiorini1998motion}. From Table \ref{tab:diffusion_rl_metrics}, we find these heuristics can reach the goal at a high rate \RALEdit{in two view modes across both domains}. \rebuttal{However, the heuristic evader} is detected by the search team much more than our methods, which leads to a lower score. The main reason is that the evader will be exposed to the cameras hidden from it, which may attract the dynamic search team and the hand-crafted evasive heuristic is not optimal for escaping. 


Although RL-only approaches (DDPG \cite{lillicrap2015continuous} \& SAC \cite{haarnoja2018soft}) directly optimize the score as the reward function, our methods outperform them on average by 108\% on the score metric across both domains. The SAC evader can be detected less than our diffusion-RL algorithm by 10.26\% in prisoner escape domain, which is interesting and a red herring. Figure \ref{fig:diff_vs_sac} shows a red colorbar for the distance between the evader and the closest agent in the pursuit team, and the diamond purple line indicates the diffusion planned global path. From the trajectory, we can see the SAC algorithm will lead the evader to the edge of the map where there is no camera but not the goal hideout. Instead, our method can loosely follow the diffusion waypoints (purple diamond) and simultaneously deviate a little to avoid the search team when they are close (indicated by magnet and orange dashed box).  In addition, our method learns to reduce the evader speed when evading, which lowers the probability of detection, but maintains a higher speed when safer. \RALEdit{We also ablate our diffusion-RL with the diffusion-only baseline (Diffuser \cite{janner2022planning}) and the 56.6\% detection rate decreasing shows the low-level RL policy is crucial to adversarial evading behaviors.}


\subsection{\RALEdit{Interpretability and Flexibility}} \label{sec:costmap_res}
As discussed in \S\ref{sec:map}, we build a costmap based on the distance between evader and pursuer which implies detection risks. The costmap serves as an interpretable and memory-efficient way to allow our hierarchical agent to plan based on past experiences \RALEdit{and will improve the flexibility in the inference stage compared with RL-only motion planning.}

Figure \ref{fig:costmap} shows the costmap, where the high-intensity region indicates the evader is close to the search team which may lead to detection. We also find that the potential detection usually happens around the cameras (see Figure \ref{fig:cam}), which is reasonable since the searching team will search the region after the evader is detected by the camera. Therefore, Figure \ref{fig:selected_path} shows the evasive agent can select the best global path which receives lowest cost \rebuttal{and} can best conceal it from the camera array. At the inference stage, we can adjust the costmap with the new information we received. For example, we can add an undesired danger zone for the evader on the costmap (see Figure \ref{fig:new_zone}) such that it will select another way to reach another hideout which will avoid the danger zone (see Figure \ref{fig:new_path}). We ablate our model with and without the costmap and show that the detection from map-based path selection is further decreased by half. 


\subsection{\RALEdit{Efficiency and Diversity in Guidance}} \label{sec:validation}

\begin{figure}[h]
    \centering
    \subfloat[A* Trajectories]{\includegraphics[width=0.21\textwidth]{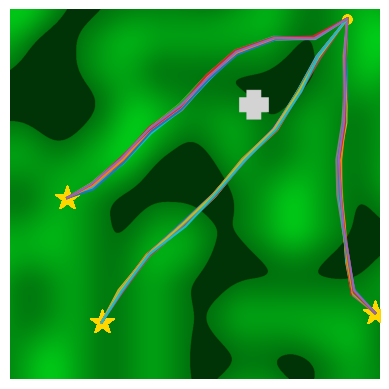} \label{fig:narrow}} 
    \hspace{0.5em}
    \subfloat[Diffusion Trajectories]{\includegraphics[width=0.21\textwidth]{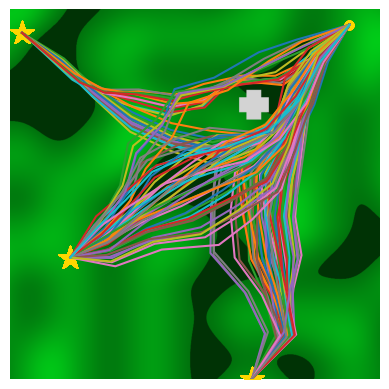} \label{fig:diverse}} 
    \vfill \vspace{0.2 em}
    \hspace{-1.3em}
    \footnotesize
    \subfloat[Inference Time (mean$\pm$std)]{
        \adjustbox{raise=3.0pc}{\begin{tabular}{lcc}
            \hline
             Traj. & RRT*           & Diffusion \\ \hline
            1          & 0.7$\pm$0.2  & 0.1$\pm$0.0  \\
            10         & 6.2$\pm$0.3  & 0.2$\pm$0.0  \\
            20         & 12.5$\pm$0.6 & 0.4$\pm$0.1  \\
            30         & 19.7$\pm$1.5 & 0.4$\pm$0.0  \\
            40         & 26.3$\pm$1.9 & 0.5$\pm$0.0  \\
            50         & 33.0$\pm$2.3 & 0.6$\pm$0.2  \\ \hline
        \end{tabular}}\label{table:diffusion_time}}
    \hspace{1em}
    \subfloat[Training Time (Hours)]{\includegraphics[width=0.2\textwidth]{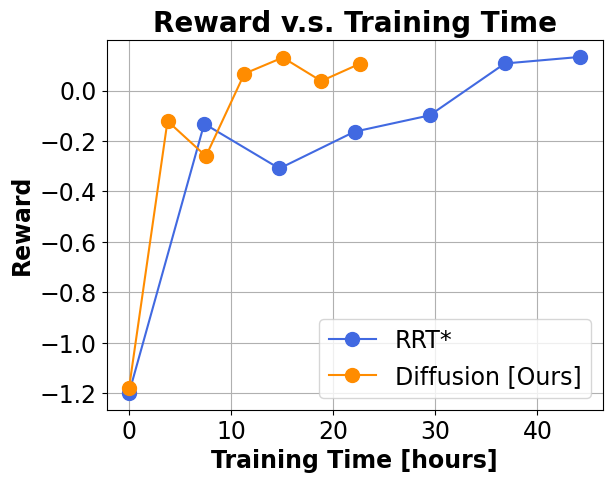} \label{fig:training_time}} 
    
    \vfill 
    \vspace{-0.0em}
    \caption{The paths from the diffusion model trained on RRT* are more diverse than A* paths (\ref{fig:narrow}-\ref{fig:diverse}). Compared to the traditional RRT* planner, the diffusion model leverages the power of parallel computing to generate trajectories an order of magnitude faster (\ref{table:diffusion_time}). \RALEdit{The diffusion-guided RL training also significantly decreases the training time (\ref{fig:training_time}).}}
    \label{fig:diffusion_traj_time}
    \vspace{-1.0em}
\end{figure}

Figure \ref{fig:narrow}-\ref{fig:diverse} compares the diffusion global paths trained on the datasets from RRT* and A*. We see the diffusion model encodes a more diverse distribution of the paths leading to the final hideouts. \RALEdit{As a result, A* guided RL method can only receive a score of 0.78$\pm$0.09, which is worse than the performance of our diffusion-RL method (0.90$\pm$0.09).} Table \ref{table:diffusion_time} shows the planning time in the inference stage for both the diffusion and RRT* algorithms. The diffusion model takes 85.7\% less time to generate one trajectory compared with RRT*, which requires a map search. Furthermore, the difference is enlarged when planning more paths since we can draw samples from the diffusion model in parallel. \RALEdit{In addition, Figure \ref{fig:training_time} shows that the diffusion-guided RL training uses less than a half time to reach the same reward level (save 25.2 hours), which greatly improves training time efficiency crucial to parameter tuning.}  We generate the results using a machine with 11th Gen Intel Core i9 processor and GeForce GTX 1660 Ti graphic card. 
\subsection{\RALEdit{Generailizability to New Camera Map}}\label{sec:generalizability}
\begin{figure}[ht]
    \vspace{-1.5em}
    \centering
    \subfloat[New Camera Map]{\includegraphics[width=0.15\textwidth, height=0.15\textwidth]{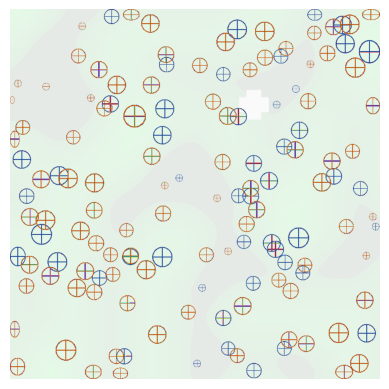} \label{fig:test_cam_dist}} 
    \subfloat[New Costmap]{\includegraphics[width=0.15\textwidth, height=0.15\textwidth]{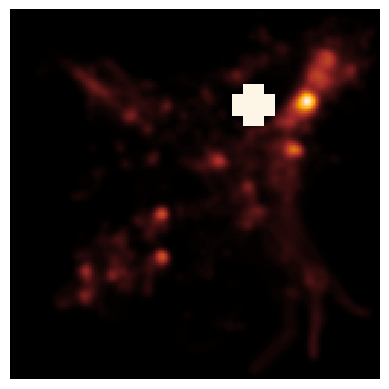}\label{fig:test_cam_heat}}
    \vspace{-0.3em}
    \caption{\RALEdit{(\ref{fig:test_cam_dist}) - Cameras are highlighted. We use a different camera distribution in the test scenario (indicated by orange) from the one in the training scenario (indicated by blue). (\ref{fig:test_cam_heat}) - we derive a new costmap reflecting the testing scenario camera distribution by running Algorithm \ref{alg:costmap}.}}
    \label{fig:generalizability}
    \vspace{-0.5em}
\end{figure}

\RALEdit{The high-level diffusion and low-level RL policies trained on one map can be generalized to other maps with different camera distributions with the only need to re-run costmap reconstruction process rather than the whole training pipeline (see Figure \ref{fig:generalizability}).}
\RALEdit{We benchmark with the new camera map, and our results are shown in the Table \ref{tab:generalizability}. Our method still outperforms the \RALEdit{best} baseline by 18.6\% \RALEdit{and outperforms 43.99\% across all baselines in average}, which shows our algorithm can be generalized to the new camera map without any retraining of diffusion or RL policies.}

\begin{table}[ht]
\vspace{-1.5em}
\centering
\caption{\RALEdit{Generalizability Benchmarks (mean $\pm$ std)}}
\vspace{-1em}
\label{tab:generalizability}
\resizebox{0.8\columnwidth}{!}{%
\begin{tabular}{lccc}
\toprule
\textbf{Domain (global-view)}                                       & \multicolumn{3}{c}{{Prisoner Escape}}              \\ \toprule
                                                      & {Score $\uparrow$} & {Detection $\downarrow$} & {Goal-Reach. $\uparrow$}  \\ \toprule
\multicolumn{1}{l|}{\textbf{Non-Learning Approaches}} &                &                    &                                     \\
\multicolumn{1}{l|}{A-Star Heuristic}                 & 0.694$\pm$0.185  & 0.376$\pm$0.229      & 0.970$\pm$0.171            \\
\multicolumn{1}{l|}{RRT-Star Heuristic}               & 0.701$\pm$0.179  & 0.367$\pm$0.220      & 0.970$\pm$0.171             \\
\multicolumn{1}{l|}{VO-Heuristic}                     & 0.728$\pm$0.104  & 0.336$\pm$0.131     & 0.980$\pm$0.140           \\
\multicolumn{1}{l|}{\textbf{Learning Approaches}}     &                &                    &                                \\
\multicolumn{1}{l|}{DDPG}                             & 0.136$\pm$0.241  & 0.847$\pm$0.264      & 0.070$\pm$0.255         \\
\multicolumn{1}{l|}{SAC}                              & 0.670$\pm$0.203  & {0.195$\pm$0.226}      & 0.130$\pm$0.336        \\
\multicolumn{1}{l|}{Diffusion Only}                   & 0.767$\pm$0.080  & 0.292$\pm$0.100      & {1.000$\pm$0.000}        \\
\multicolumn{1}{l|}{\textbf{Our Approaches}}          &                &                    &                       \\
\multicolumn{1}{l|}{Diffusion-RL [ours]}                     & 0.864$\pm$0.111  & 0.155$\pm$0.116      & 0.940$\pm$0.237    \\
\multicolumn{1}{l|}{Diffusion-RL-Map [ours]}                 & \textbf{0.910$\pm$0.075}  & 0.113$\pm$0.093      & 1.000$\pm$0.000 \\ \toprule
\end{tabular}
}
\vspace{-1em}
\end{table}

\subsection{Real Robot Demonstration}\label{sec:robotarium}
\begin{figure}[h]
    \centering
    \subfloat[RRT* Heuristic]{\includegraphics[width=0.15\textwidth]{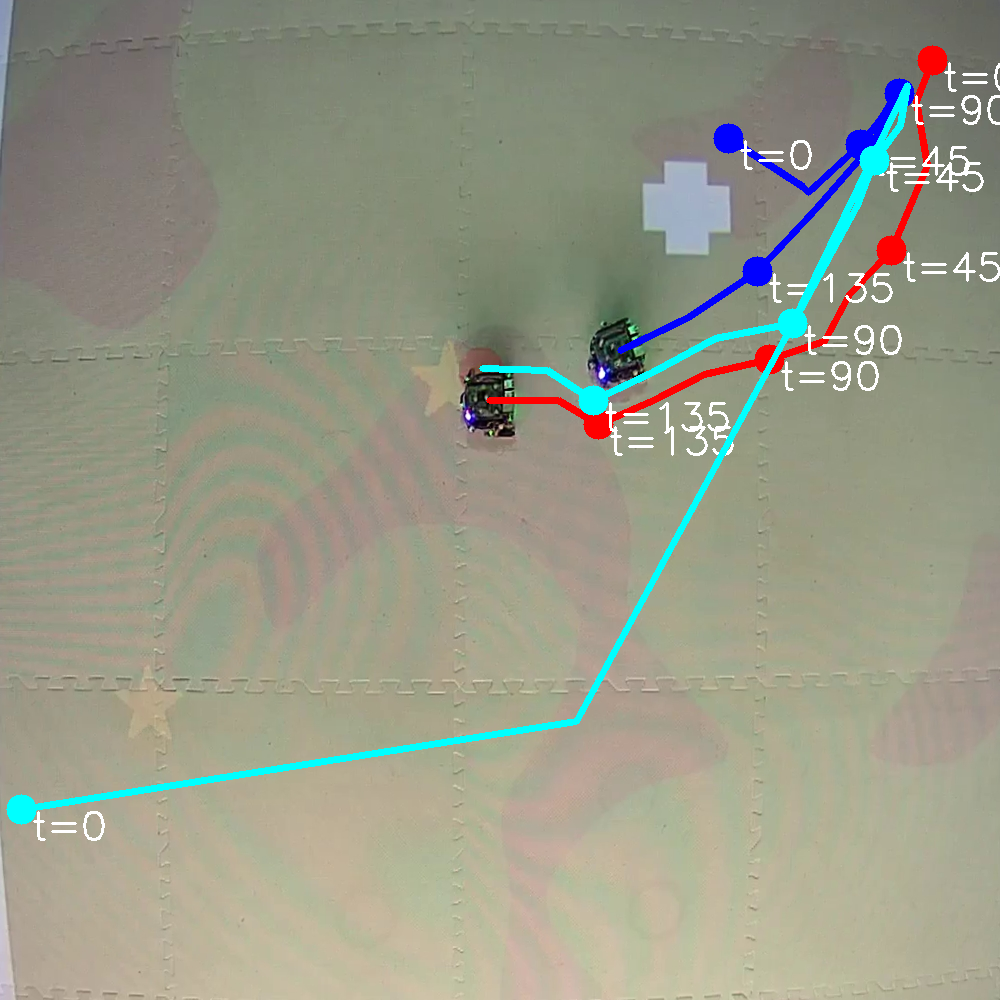} \label{fig:rrt_star}} 
    \subfloat[\RALEdit{SAC}]{\includegraphics[width=0.15\textwidth]{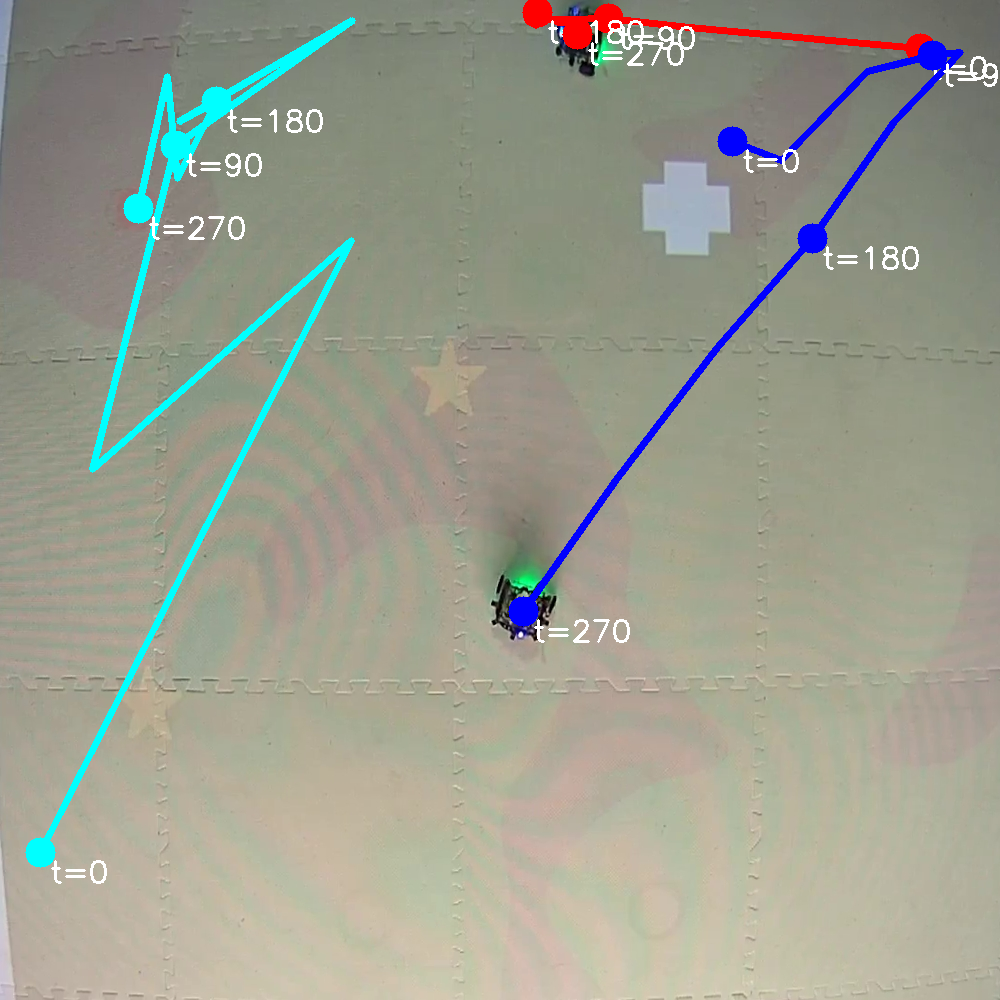} \label{fig:diffusion_rl}} 
    \subfloat[Diffusion-RL-Map]{\includegraphics[width=0.15\textwidth]{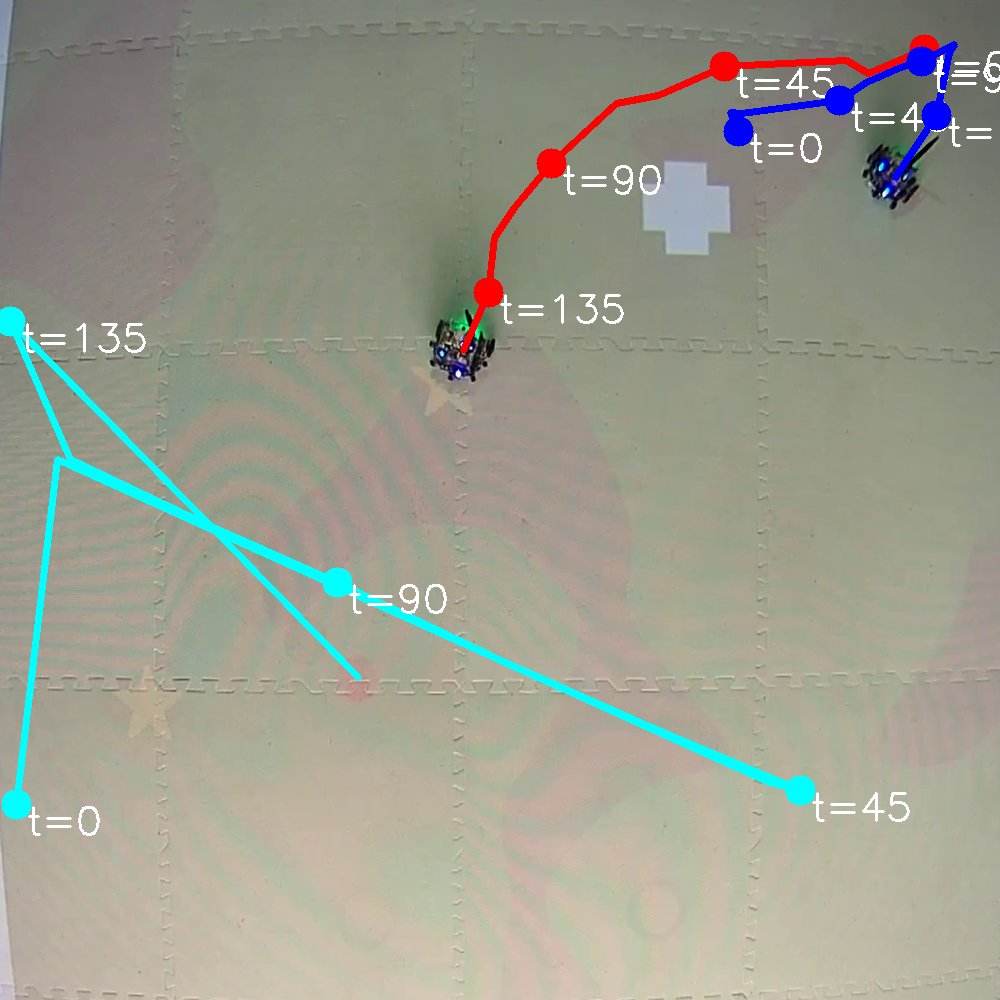} \label{fig:sac}}
    \vspace{-0.5em}
    \caption{Pursuit-evasion trajectories: the evader, search party, and helicopter trajectory are indicated with red, blue, and cyan, respectively. We use a diffusion model to guide the evader to the hideout and explore escaping behaviors with RL, which performs better than RRT* heuristic with insufficient evasive behaviors and SAC method converging to a suboptimal place.}
    \label{fig:robotarium}
    \vspace{-0.5em}
\end{figure}

We also test our method with the Robotarium \cite{8960572} testbed to validate our method with real robot dynamics. Figure \ref{fig:robotarium} shows the comparison among the RRT* heuristic, \RALEdit{SAC}, and our method. \RALEdit{We find the evader (red trajectory) equipped with our policy can reach the hideout and remain untracked by the searching team (blue and cyan trajectories). In comparison, the SAC policy does not lead the evader to the hideout, and the RRT* heuristic travels a path near cameras and is tracked by the helicopter starting at $t\approx90$.} This validates our benchmarking results and indicates that our hierarchical data-driven learning-based method can outperform the baselines from the efficient diffusion-guided exploration, RL-based trial-and-error multi-agent interaction, and map optimizations. 

\subsection{Discussion}\label{sec:discussion}
The significant disparity in goal-reaching rates between standalone RL and our approaches indicates that RL struggles with long-horizon exploration in large partially observable domains, supporting our premise in \S\ref{sec:Intro} that diffusion models can guide RL exploration. Additionally, Figure \ref{fig:diff_vs_sac} reveals that SAC can become trapped in a local minimum. From this, we infer that our methods can achieve long-term planning by providing the RL agent with a diffusion-based global plan while retaining the agility to evade opponent agents in the short horizon. \RALEdit{The ablation studies in \S\ref{sec:costmap_res} to \S\ref{sec:generalizability}}  demonstrate that our \RALEdit{hierarchy can generate more interpretable paths, have more flexible motion plans and generalize to different test maps. Importantly, it significantly improves training and inference efficiency, which unlocks \rebuttal{the} potential on more applications in real-time tasks \cite{wu2024learningwheelchairtennisnavigation, lee2024learning}. }


%% file: conc.tex


\section{Conclusion and Future Work}
In conclusion, we have presented a novel learning-based framework for evaders in large, partially observable, multi-agent pursuit-evasion settings. 
Our approach can help the evader reach the hideout and best conceal itself from capturing by using a hierarchy with a high-level diffusion path planner and a low-level RL evasive policy. 
In the future, we plan to train the whole framework end-to-end and bring our diffusion-RL technique to \rebuttal{real-world} navigation tasks involving high-level human domain knowledge transfer and low-level dynamic constraints (e.g. environment-aware path generation and \rebuttal{finetuning} in autonomous driving).


